%% file: main.tex
\documentclass[fleqn,10pt]{wlscirep}
\usepackage[utf8]{inputenc}
\usepackage[T1]{fontenc}
\usepackage{tabularx}
\usepackage{array,  makecell}
\usepackage{tikz}
\usetikzlibrary{bayesnet}
\newcommand{\PreserveBackslash}[1]{\let\temp=\\#1\let\\=\temp}
\newcolumntype{C}[1]{>{\PreserveBackslash\centering}p{#1}}
\newcolumntype{R}[1]{>{\PreserveBackslash\raggedleft}p{#1}}
\newcolumntype{L}[1]{>{\PreserveBackslash\raggedright}p{#1}}
\usepackage[export]{adjustbox}
\usepackage{layouts}
\usepackage{float}
\usepackage{longtable}

\title{Using cognitive psychology to understand GPT-3}

\author[1,*]{Marcel Binz}
\author[1]{Eric Schulz}
\affil[1]{MPRG Computational Principles of Intelligence, Max Planck Institute for Biological Cybernetics, 72076 T\"ubingen, Germany}

\affil[*]{marcel.binz@tue.mpg.de}

\begin{abstract}
We study GPT-3, a recent large language model, using tools from cognitive psychology. More specifically, we assess GPT-3's decision-making, information search, deliberation, and causal reasoning abilities on a battery of canonical experiments from the literature. We find that much of GPT-3's behavior is impressive: it solves vignette-based tasks similarly or better than human subjects, is able to make decent decisions from descriptions, outperforms humans in a multi-armed bandit task, and shows signatures of model-based reinforcement learning. Yet we also find that small perturbations to vignette-based tasks can lead GPT-3 vastly astray, that it shows no signatures of directed exploration, and that it fails miserably in causal reasoning task. These results enrich our understanding of current large language models and pave the way for future investigations using tools from cognitive psychology to study increasingly capable and opaque artificial agents.
\end{abstract}
\begin{document}

\flushbottom
\maketitle

\section*{Introduction}

% general things to mention: we let GPT-3 
% looking at task performance is not enough – compare to humans
With the advent of increasingly capable artificial agents, comes the urgency to improve our understanding of how they learn and make decisions \cite{gunning2019xai}. Take as an example large language models \cite{brown2020language}. These models' abilities are, by many standards, impressive. They can generate text that human evaluators have difficulty distinguishing from text written by other humans \cite{brown2020language}, generate computer code \cite{chen2021evaluating}, or converse with humans about a range of different topics \cite{lin2020caire}. What is perhaps even more surprising, is that these models' abilities go beyond mere language generation: they can, for instance, also play chess at a reasonable level \cite{noever2020chess} and solve university-level math problems \cite{drori2021neural}. These observations have prompted some to argue that this new class of \emph{foundation models}, which are models trained on broad data at scale and adapted to a wide range of downstream tasks, shows some form of general intelligence \cite{chalmers2020gpt}. Yet others have been more skeptical, pointing out that these models are still a far cry away from a human-level understanding of language and semantics \cite{marcus2020gpt}. But how can we evaluate whether or not these models --at least in some situations-- learn and think like people? One approach towards evaluating a model's human-likeness comes from cognitive psychology. Psychologists, after all, are experienced in trying to formally understand another notoriously impenetrable algorithm: the human mind. 

In the present article, we investigate the Generative Pre-trained Transformer 3 model (or short: GPT-3) \cite{brown2020language} on several experiments taken from the cognitive psychology literature. Our analyses cover two types of experiments: vignette-based and task-based experiments. While vignette-based experiments involve a short and predefined description of a hypothetical scenario, task-based experiments are programmatically generated on a trail-by-trial basis. The selected tasks for both of these settings cover well-known areas of cognitive psychology: decision-making, information search, deliberation, and causal reasoning. We are primarily interested in whether GPT-3 can solve these tasks appropriately as well as how its behavior compares to human subjects. Our results show that GPT-3 can solve challenging vignette-based problems. Yet, we also highlight that these vignettes or similar texts might have been part of its training set. Moreover, we find that GPT-3's behavior strongly depends on how the vignettes are presented. Thus, we also subject GPT-3 to a battery of task-based problems. The results from these task-based assessments show that GPT-3 can make human-level decisions in both description-based and experience-based decision-making experiments, yet does not learn and explore in a human-like fashion. Furthermore, even though GPT-3 shows signatures of model-based reinforcement learning, it fails altogether in a causal reasoning task. Taken together, our results improve our understanding of current large language models, suggest ways in which they can be improved, and pave the way for future investigations using tools from cognitive psychology to study increasingly capable and opaque artificial agents.

\section*{GPT-3}

GPT-3 is an auto-regressive language model \cite{brown2020language}. It utilizes the transformer architecture \cite{vaswani2017attention} --a deep learning model that heavily relies on the mechanism of self-attention-- to produce human-like text. Just like recurrent neural networks, transformers are designed to process sequential data, such as natural language. However, unlike recurrent neural networks, transformers process the entire data all at once, with the attention mechanism providing context for any position in the input sequence. The model itself is large, it has 175 billion parameters, and it was trained on a vast amount of text: hundreds of billions of words from the internet and books. GPT-3's architecture is similar to that of its predecessor, GPT-2 \cite{radford2018improving}, but contains many more trainable parameters. Thus, GPT-3 can be thought of as an experiment in massively scaling up known algorithms \cite{sutton2019bitter}. Larger models can capture more of the complexities of the data they were trained on and can transfer this knowledge to tasks that they have not been specifically trained for. Rather than being fine-tuned on a problem, these large language models can be given an instruction together with some examples of the task and identify what to do based on this alone. This is called ``in-context learning'' because the model picks up on patterns in its ``context'', for example, the string of words that the model is asked to complete. GPT-3 does incredibly well at in-context learning across a range of settings \cite{liu2021makes}, sometimes even performing at a level comparable to the best fine-tuned models \cite{brown2020language,lampinen2022can}. Since GPT-3 is one of the biggest and most versatile large language models, it is a good candidate to be scrutinized using cognitive psychology.

\section*{A cognitive psychology view on GPT-3}

We will subject GPT-3 to several tasks taken from the cognitive psychology literature. These tasks fall into four categories: 1. decision-making, 2. information search, 3. deliberation, and 4. causal reasoning. We will begin our investigations with several, classical vignette-based problems. For these vignette-based investigations, we confronted GPT-3 with text-based descriptions of hypothetical situations while collecting its responses. However, as we will point out, these vignettes have the problem that GPT-3 has likely experienced identical or similar such tasks in its training data. Moreover, we found that GPT-3's response can be tampered with just by marginally changing the vignettes and thereby creating adversarial situations. Thus, we also evaluated GPT-3's abilities in various task-based experiments. In these task-based investigations, we take canonical tasks from the literature and emulate their experimental structure as programmatically generated text to which GPT-3 responds on every experimental trial. We then use GPT-3's responses to analyze its behavior similar to how cognitive psychologists would analyze human behavior in the same tasks.

\section*{Results}

\begin{figure}[t]
    \centering

    \begin{tabular}{@{}ll}
        \textbf{A} & \textbf{B} \\
        \raisebox{0.cm}{\begin{tikzpicture}
        \node[draw=black!70, inner sep=10pt, rounded corners=0.15cm, text width=6.6cm, minimum height=5cm, align=left] (prompt) {
Linda is 31 years old, single, outspoken, and very bright. She majored in philosophy. As a student, she was deeply concerned with issues of discrimination and social justice, and also participated in anti-nuclear demonstrations. \\[0.5cm]

Q: Which option is the most probable?\\[0.5cm]
- Option 1: Linda is a bank teller.\newline
- Option 2: Linda is a bank teller and is active in the feminist movement.\newline
- Option 3: Linda is a member of the NRA.\\[0.5cm]
A: Option};
    \end{tikzpicture}} &
    \includegraphics[scale=0.37]{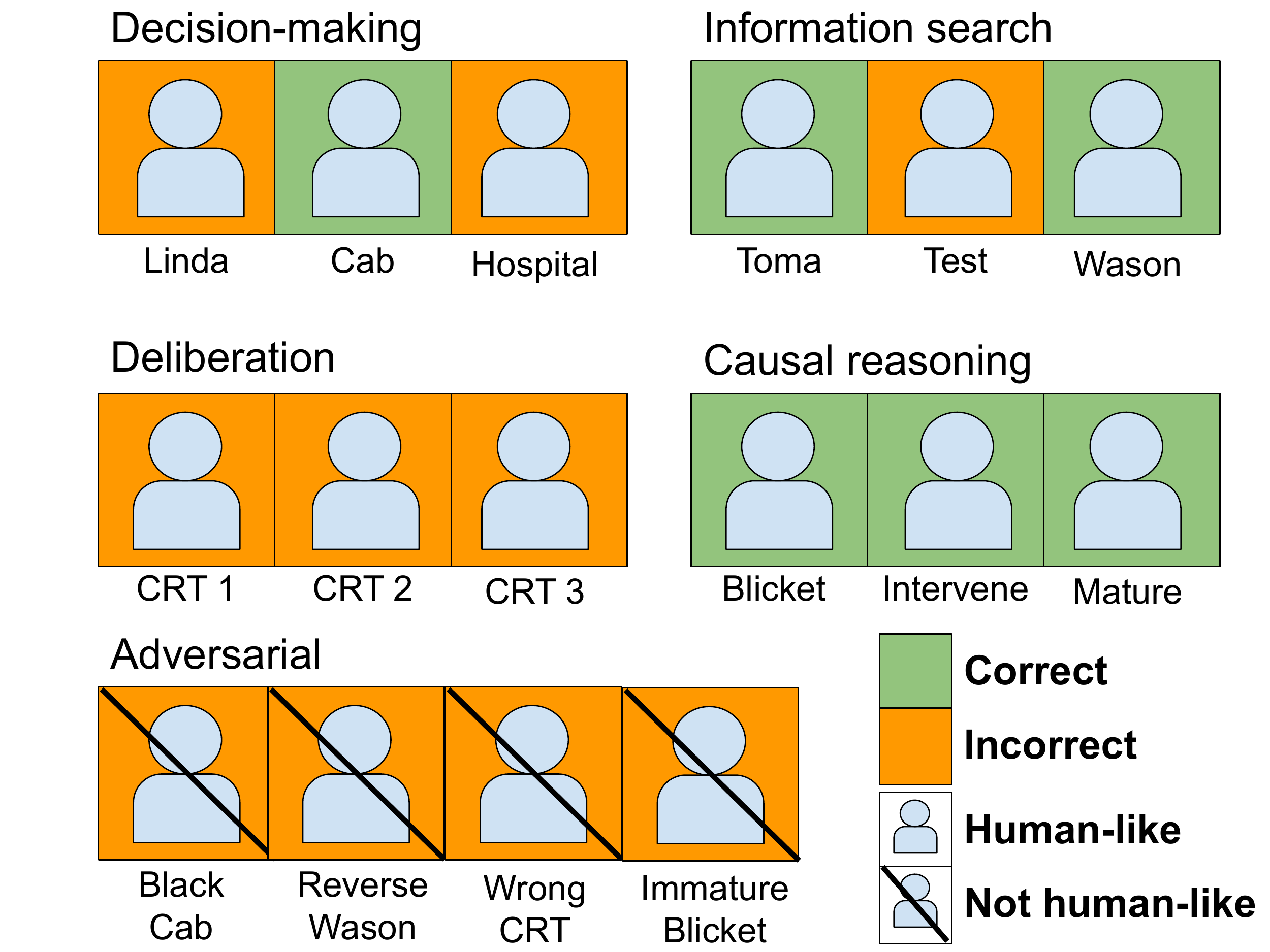} 
    \end{tabular}
    \caption{Vignette-based tasks. \textbf{A:} Example prompt of hypothetical scenario, in this case the famous Linda problem, as submitted to GPT-3. \textbf{B:} Results. While in 12 out 12 standard vignettes, GPT-3 either answers correctly or makes human-like mistakes, it makes mistakes that are not human-like when given the adversarial vignettes.}
    \label{fig:vignettes}
\end{figure}
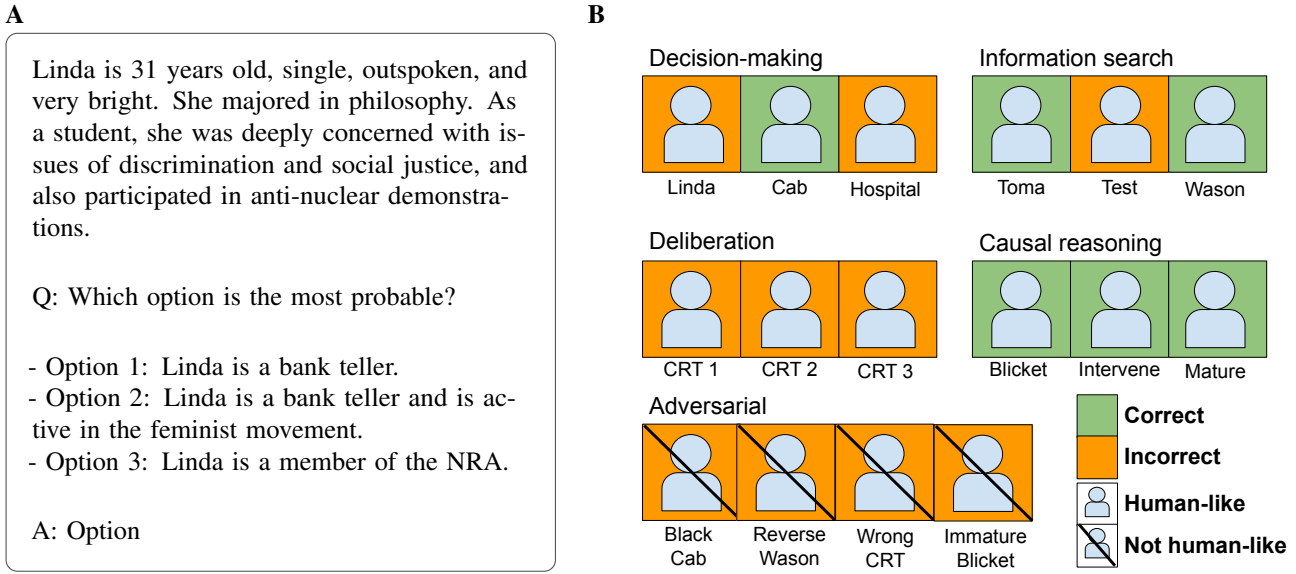

We used the public OpenAI API to run all our simulations \cite{OpenAIAPI}. There are four GPT-3 models accessible through this API: ``Ada'', ``Babbage'', ``Curie'' and ``Davinci'' (sorted from the least to the most complex model). We focused our investigation on the most powerful of these models (``Davinci'') unless otherwise noted. We furthermore set the temperature parameter to $0$, leading to deterministic answers, and kept the default values for all other parameters.

\subsection*{Vignette-based investigations}
For the vignette-based investigations, we took canonical scenarios from the cognitive psychology literature, entered them as prompts into GPT-3, and recorded its answer. For each scenario, we report if GPT-3 responded correctly or not. Moreover, we classified each response as something a human could have said because it was either the correct response or a mistake commonly observed in human data. For cases where there were only two options, one correct and one that is normally chosen by human subjects, we added a third option that was neither correct nor plausibly chosen by people. The following subsections briefly summarize our main findings. We refer the reader to SI Appendix for a detailed description of the submitted prompts and GPT-3's corresponding answers.

\subsubsection*{Decision-making: Heuristics and biases}
We began our investigations of GPT-3's decision-making by prompting the canonical ``Linda problem'' \cite{tversky1983extensional} (Linda, see Figure \ref{fig:vignettes}A). This problem has been known to assess the conjunction fallacy, a formal fallacy that occurs when it is assumed that specific conditions are more probable than a single general one. In the standard vignette, a hypothetical woman named Linda is described as ``outspoken, bright, and politically active''. Participants are then asked if it was more likely that Linda is a bank teller or that she is a bank teller \emph{and} an active feminist. GPT-3, just like people, chose the second option, thereby falling for the conjunction fallacy.

Next, we prompted the so-called ``cab problem'' \cite{tversky2015causal} (Cab, see SI Appendix), in which participants commonly fail to take the base rate of different colors of taxis in a city into account when judging the probability of the color of a cab that was involved in an accident. Unlike people, GPT-3 did not fall for the base-rate fallacy, i.e. to ignore the base rates of different colors, but instead provided the (approximately) correct answer.

Finally, we asked GPT-3 to provide an answer to the ``hospital problem'' \cite{kahneman1972subjective} (Hospital, see SI Appendix), in which participants are asked which of two hospitals, a smaller or a larger one, is more likely to report more days on which more than 60\% of all born children were boys. While the correct answer would be the smaller hospital (due to the larger variance of smaller samples), GPT-3, just like people, thought that the probability was about equal.

\subsubsection*{Information search: Questions and hypothesis tests}
%We wanted to see how GPT-3 searches for information, which is a key feature of human behavior \cite{coenen2019asking}. 

First, we assessed if GPT-3 can adaptively change between constraint-seeking vs. hypothesis-scanning questions. Constraint-seeking questions target a feature shared by multiple objects, such as ``Is the person female?'', whereas hypothesis-scanning questions target a single object, such as ``Is the person Linda?''. Crucially, which type of question is more informative depends on past observations. Ruggeri et al. \cite{ruggeri2017toma} manipulated the particular reasons for why a fictitious character named Toma was repeatedly late to school (Toma, see SI Appendix). While for one group he was frequently late because his bicycle had broken, for the other group he was late for various reasons with half of them being that he could not find various objects. While trying to find out why Toma is late to school again, the first group should ask the hypothesis-scanning question  ``Was he late because his bicycle broke?'', whereas the second group should ask the constraint-seeking question ``Was he late because he could not find something?''. GPT-3 picked the appropriate question in each scenario.

Secondly, we confronted GPT-3 with a scenario originally presented by Baron et al. \cite{baron1988heuristics} in which subjects need to choose an appropriate test to discriminate between two illnesses (Test, see SI Appendix). Empirically, participants tend to choose the wrong test, likely because they overvalue questions that have a high probability of a positive result given the most likely hypothesis. GPT-3, just like people, fell for the same congruence bias.

Finally, we presented Wason's well-known ``Card Selection Task'' \cite{wason1968reasoning} to GPT-3, explaining that the visible faces of four cards showed A, K, $4$ and $7$, and that the truth of the proposition ``If a card shows a vowel on one face, then its opposite face shows an even number'' needed to be tested (Wason, see SI Appendix). GPT-3 suggested to turn around A and $7$, which is commonly accepted as the correct answer, even though most people turn around A and $4$.

\subsubsection*{Deliberation: The Cognitive Reflection Test}

We also tried to estimate GPT-3's tendency to override an incorrect fast response with answers derived by further deliberation. For this, we prompted the three items of the Cognitive Reflection Test \cite{frederick2005cognitive} (CRT1-CRT3, see SI Appendix). One example item of this task is: ``If it takes $5$ machines $5$ minutes to make $5$ widgets, how long would it take $100$ machines to make $100$ widgets?''. While the initial response might be to say ``$100$'', $100$ machines would just be as fast as $5$ machines and thus also take $5$ minutes. For all three items of the CRT, GPT-3 responded with the intuitive but incorrect answer, as has been observed in earlier work \cite{nye2021improving}.

\subsubsection*{Causal reasoning: Blickets, interventions, and counterfactuals}

We lastly assessed GPT-3's causal reasoning abilities. In a first test, we prompted GPT-3 with a version of the well-known ``Blicket'' experiment \cite{sobel2007blicket} (Blicket, see SI Appendix). For this, blickets are introduced as objects that turn on a machine. Afterward, two objects are introduced. The first object turns on the machine on its own. The second machine does not turn on the machine on its own. Finally, both objects together turn on the machine. GPT-3, just like people, managed to correctly identify that the first but not the second object is a blicket.

In a second test, we asked GPT-3 to intervene in a scenario by removing the correct object to prevent an effect after having read about three different objects, one causing and two not causing the effect (in this case, an allergic reaction; Intervene, see SI Appendix). GPT-3 identified the correct object to be removed.

In the final test, we probed GPT-3's ability of mature causal reasoning \cite{nyhout2019mature} (Mature, see SI Appendix). In this task, GPT-3 was told that there were four pills: A, B, C and D. While A and B individually could kill someone, C and D could not. GPT-3 successfully answered multiple questions about counterfactuals correctly, such as: ``A man took pill B and pill C and he died. If he had not taken pill B, could he still have died?''. 

\subsubsection*{Problems with vignette-based investigations}

Of the $12$ vignette-based problems presented to GPT-3, it answered six correctly and all $12$ in a way that could be described as human-like (Figure \ref{fig:vignettes}B). Does this mean that GPT-3 could pass as a human in a cognitive psychology experiment? We believe that the answer, based on the vignette-based tasks alone, has to be ``No.''. Since many of the prompted scenarios were taken from famous psychological experiments, there is a chance that GPT-3 has encountered these scenarios or similar ones in its training set. Moreover, in additional investigations, we found that many of the vignettes could be slightly modified, i.e., made into adversarial vignettes, such that GPT-3 would give vastly different responses. In the cab problem, for example, it is clearly stated that $15$\% of the cabs are blue and $85$\% are green. Yet asking GPT-3 about the probability that a cab involved in an accident was black, it responded with ``$20$\%'' (Black Cab, see SI Appendix). Simply changing the order of the options in Wason's card selection task from ``A, K, $4$, and $7$'' to ``$4$, $7$, A, and K'' caused GPT-3 to suggest turning around ``A'' and ``K'' (Reverse Wason, see SI Appendix). Giving GPT-3 the first item of the CRT and stating that ``The bat costs \$$1.00$ more than the bat.'', it still thought that the ball was \$$0.10$ (Wrong CRT, see SI Appendix). Finally, when phrasing the mature causal reasoning problem as a ``Blicket'' problem in which machines could be turned on or off, GPT-3 answered some questions incorrectly while contradicting itself in its explanations (Immature Blicket, see SI Appendix). There have recently been other, much larger investigations using similar vignettes, whose results agree largely with our assessment \cite{bigbench2022}.

% I modified this a bit and moved it below
%Given these results, it is clear that GPT-3 can produce passable results on some vignette-based tasks, but that whether it is just behaving like a parrot or --at least at times-- reasoning successfully cannot be sufficiently decided based on vignettes alone. We therefore turned the lens of our investigations to task-based assessments of GPT-3's abilities.

%%%%%%%%%%%%%%%%%%%%%%%%%%%%%%%%%%%%%%%%%%%%%%%%%%%%%%%%%%%%%%%%%%%%%%%%%%%%%%%%%
%TASK BASED
%%%%%%%%%%%%%%%%%%%%%%%%%%%%%%%%%%%%%%%%%%%%%%%%%%%%%%%%%%%%%%%%%%%%%%%%%%%%%%%%%
\subsection*{Task-based investigations}

The results from the previous section indicate that GPT-3 can produce passable responses in some vignette-based tasks. It is, however, not possible to decide whether it is merely behaving like a parrot, repeating what it has seen in the training data, or whether it is reasoning successfully. We, therefore, next turned our lens of investigation to a more challenging setting and tested GPT-3 on actual, task-based experiments. In order to do so, we selected a set of four classical experiments that we believe to be representative of the cognitive psychology literature. For each of these, we programmatically generated a description that was entered as a prompt and --if there were multiple trials-- updated the text with GPT-3's response and the received feedback. 

%We subjected GPT-3 to various task-based investigations. For this, we generated the descriptions that were entered as a prompt programatically, and --if there were multiple trials-- updated the text with GPT-3's previous responses. We then extracted GPT-3's responses and analyzed them just how cognitive psychologists would analyze human behavior. We again set GPT-3's temperature parameter to 0 and --unless stated otherwise-- focused on the ``Davinci'' engine's behavior.

\subsubsection*{Decision-making: Decisions from descriptions}

\begin{figure}[ht!]
    \begin{flushleft}
    \begin{tabular}{@{}ll}
    \textbf{A} & \textbf{B} \\
    \raisebox{0.6cm}{\begin{tikzpicture}
        \node[draw=black!70, inner sep=0.2in, rounded corners=0.15cm, text width=3.in, minimum height=4cm, align=left] (prompt) {Q: Which option do you prefer? \\[0.5cm] - Option F: $69.0$ dollars with $1.0$\% chance, $26.0$ dollars with $99.0$\% chance. \\ - Option J: $2.0$ dollars with $75.0$\% chance, $94.0$ dollars with $25.0$\% chance. \\[0.5cm] A: Option};
    \end{tikzpicture}} & \includegraphics[]{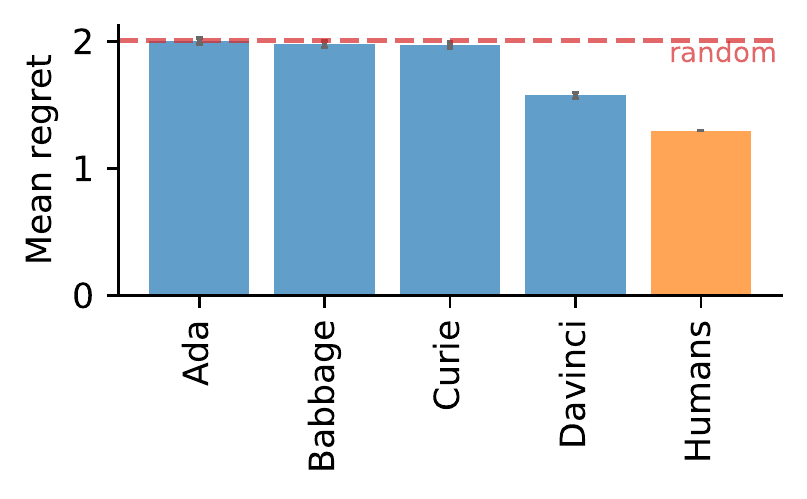} \end{tabular} \\ \vspace{0.2cm}
    
    \begin{tabular}{@{}l}
        \textbf{C} \\
        \includegraphics[]{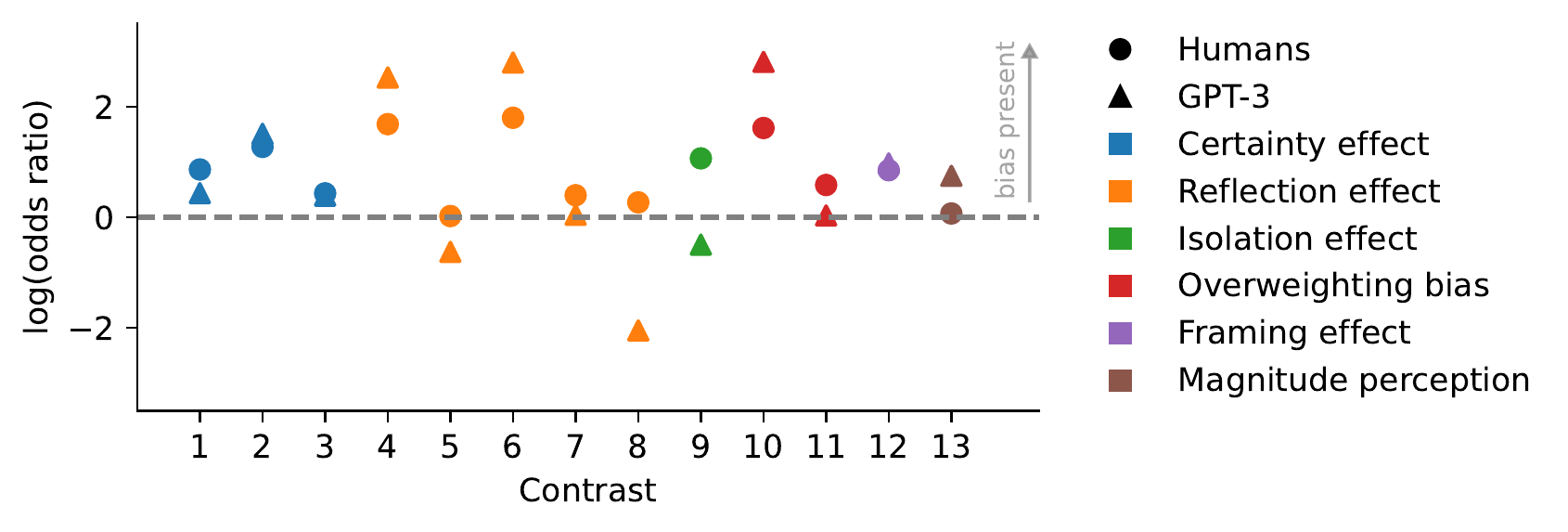} 
    \end{tabular}
    \end{flushleft}
    \centering
    \caption{Decisions from descriptions. \textbf{A:} Example prompt of a problem provided to GPT-3. \textbf{B:} Mean regret averaged over all 13,000 problems taken from Peterson et al. \cite{peterson2021using}. Lower regret means better performance. Error bars indicate the standard error of the mean. \textbf{C:} Log-odds ratios of different contrasts used to test for cognitive biases. Positive values indicate that the given bias is present in humans (circle) or GPT-3 (triangle). Human data adapted from Ruggeri et al. \cite{ruggeri2020replicating}. For a detailed description of this analysis, see SI Appendix.}
    \label{fig:dfd}
\end{figure}
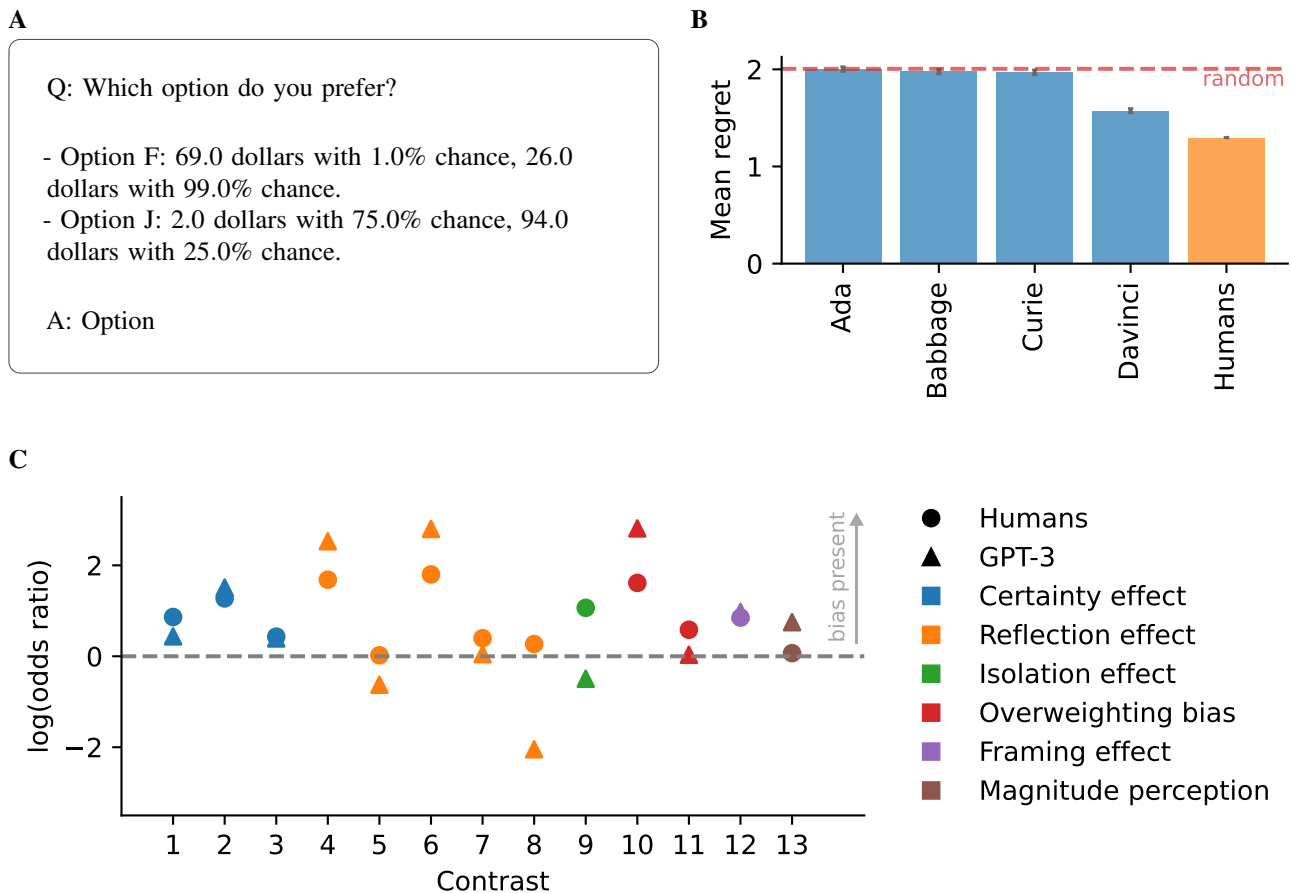

How people make decisions from descriptions is one of the most well-studied areas of cognitive psychology, ranging from the early, seminal work of Kahneman \& Tversky \cite{kahneman1979prospect} to modern, large-scale investigations \cite{ruggeri2020replicating, peterson2021using}. In the decisions from descriptions paradigm, a decision-maker is asked to choose between one of two hypothetical gambles like the ones shown in Figure~\ref{fig:dfd}A. To test whether GPT-3 can reliably solve such problems, we presented the model with over $13,000$ problems taken from a recent benchmark data-set \cite{peterson2021using}. Figure \ref{fig:dfd}B shows the regret, which is defined as the difference between the expected outcome of the optimal option and that of the actually chosen option, obtained by different models in the GPT-3 family and compares their performance to human decisions. We found that only the largest of the GPT-3 models (``Davinci'') was able to solve these problems above chance-level ($t(29134) = -16.85, p = < .001$), whereas the three smaller models did not (all $p > 0.05$). While the ``Davinci'' model did reasonably well, it did not reach human-level performance ($t(29134) = -11.50, p < .001$). 

%how people make decisions from descriptions is one of the most well-studied areas of psychology, ranging from the early investigations of Kahneman \& Tversky \cite{kahneman1979prospect} to recent large-scale investigations \cite{ruggeri2020replicating, peterson2021using}. In the more general setting, a decision-maker is asked to choose between one of two hypothetical gambles. We were interested in whether GPT-3 could reliably solve decisions from descriptions-problems like the one illustrated in Figure~\ref{fig:dfd}A. To test this, we presented the model with over $13,000$ problems taken from a recent benchmark data-set \cite{peterson2021using}. Figure \ref{fig:dfd}B shows the amount of money obtained by different models in the GPT-3 family, and compares their performance to human decisions. We found that the largest of the GPT-3 models (i.e. ``Davinci'') was the only model that solved these problems above chance-level, $t(4000) = 2.3, p = .011$. The three smaller models did not exceed chance-level performance (all $p>0.05$). Therefore, we focused our main analyses on the ``Davinci'' model. While the ``Davinci'' model did reasonably well at the investigated problems, it did not reach human-level performance, $t(4000) = 2.3, p = .011$. 

However, given that GPT-3 was not too far away from human performance, it is reasonable to ask whether the model also exhibited human-like, cognitive biases. In their original work on prospect theory, Kahneman \& Tversky \cite{kahneman1972subjective} identified several biases of human decision-making by contrasting answers to multiple carefully selected problems pairs. We replicated the original analysis of Kahneman \& Tversky using choice probabilities of GPT-3 and found that GPT-3 showed three of the six biases identified by  Kahneman \& Tversky. First, it displayed a framing effect, meaning that its preferences changed depending on whether a choice was presented in terms of gains or losses. GPT-3 was also subject to a certainty effect, meaning that it preferred guaranteed outcomes to risky ones even when they had slightly lower expected values. Finally, GPT-3 showed an overweighting bias and assigned higher importance to a difference between two small probabilities (e.g., $1\%$ and $2\%$) than to the same differences between two larger probabilities (e.g., $41\%$ and $42\%$). Figure \ref{fig:dfd}C contains an analysis of these three biases and the three additional ones we did not find in GPT-3. For a detailed description of the conducted analysis, see SI Appendix.

%To illustrate this, consider the following two problems: \\

%\noindent\textbf{Problem 1:} \\
%\noindent\text{Imagine we gave you 1,000 right now to play a game. Which option would you prefer?} \\
%\text{- Option F: A $50\%$ chance to gain an additional $1,000$ ($50\%$ chance of gaining $0$ beyond what you already have).} \\
%\text{- Option J: A $100\%$ guarantee of gaining an additional $500$.} \\

%\noindent \textbf{Problem 2:} \\
%\noindent\text{Imagine we gave you 2,000 right now to play a game. Which option would you prefer?} \\
%\text{- Option F: A $50\%$ chance you will lose $1,000$ ($50\%$ chance of losing $0$).} \\
%\text{- Option J: A $100\%$ chance you will lose $500$.} \\

%The two options in both of these problems are identical, but humans and GPT-3 tend to select option J in the first problem and option F in the second. 
% mention that is it impressive that the models give valid answers in the first place?

\subsubsection*{Information search: Directed and Random Exploration}

GPT-3 did well in the vignette-based information search tasks, so we were curious how it would fare in a more complex setting. The multi-armed bandit paradigm provides a suitable test-bed for this purpose. It extends the decisions from descriptions paradigm from the last section by adding two layers of complexity. First, the decision-maker is not provided with descriptions for each option anymore but has to learn their value from noisy samples, i.e. from experience \cite{hertwig2004decisions}. Second, the interaction is not confined to a single choice but potentially involves repeated decisions about which option to sample. Together, these two modifications call for an important change in how a decision-maker must approach such problems. It is not enough to merely exploit currently available knowledge anymore, but also crucial to explore options that are unfamiliar and thereby gain information about their value. Previous research suggests that people solve this exploration-exploitation trade-off by applying a combination of two distinct strategies: directed and random exploration \cite{wilson2014humans}. Whereas directed exploration encourages the decision-maker to collect samples from previously unexplored options, random exploration strategies inject some form of stochasticity into the decision process \cite{gershman2018deconstructing,schulz2019algorithmic}.

%But how would it behave if it needed to trade off between seeking out information and choosing options that it knew to be good?

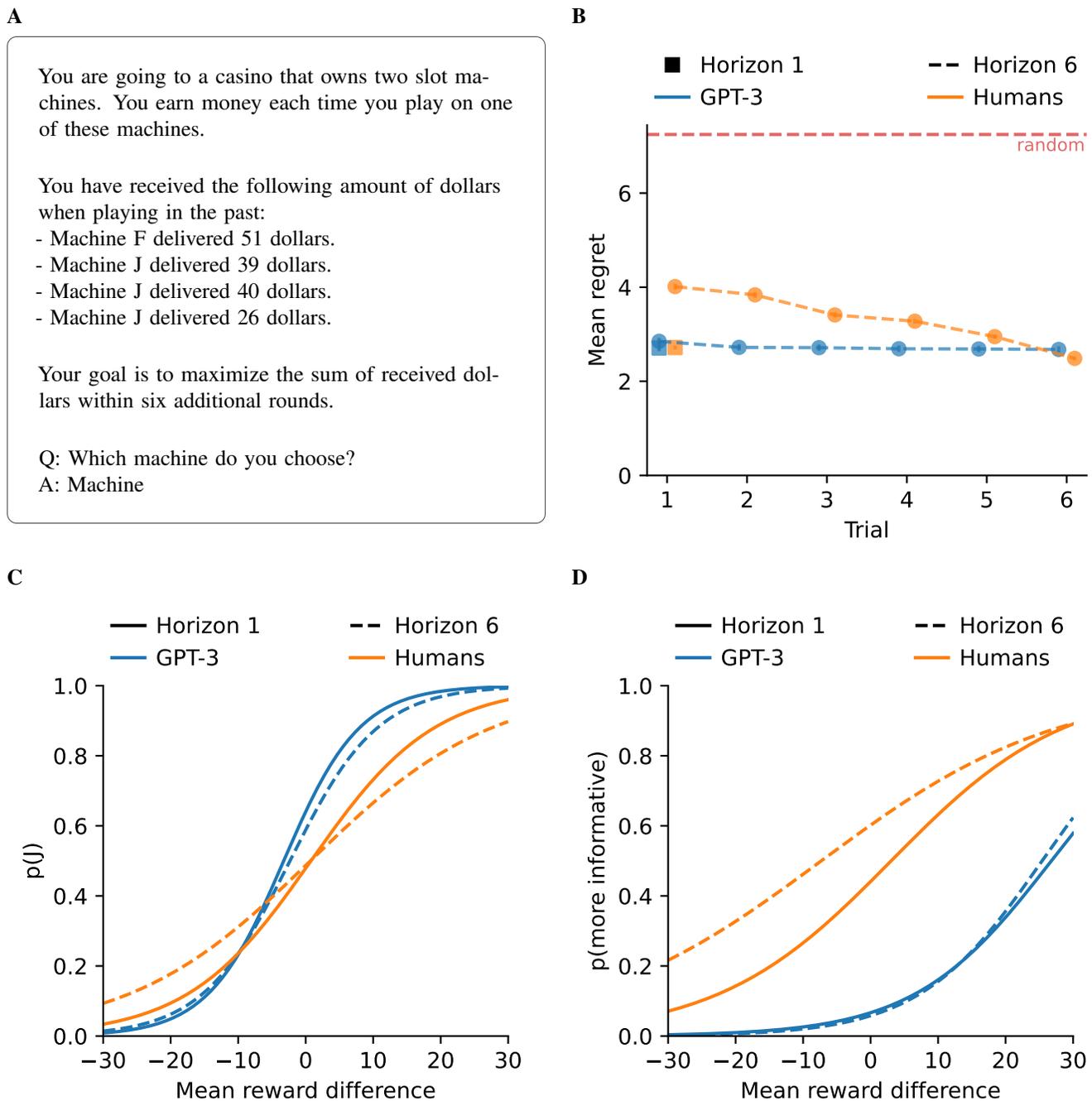
\begin{figure}[ht!]
    \centering

    \begin{tabular}{@{}ll}
        \textbf{A} & \textbf{B} \\
        \raisebox{0.55cm}{\begin{tikzpicture}
        \node[draw=black!70, inner sep=0.2in, rounded corners=0.15cm, text width=3.in, minimum height=5cm, align=left] (prompt) {You are going to a casino that owns two slot machines. You earn money each time you play on one of these machines. \\[0.5cm]
        
        You have received the following amount of dollars when playing in the past: \\
        
- Machine F delivered 51 dollars. \\
- Machine J delivered 39 dollars. \\
- Machine J delivered 40 dollars. \\
- Machine J delivered 26 dollars. \\[0.5cm]

Your goal is to maximize the sum of received dollars within six additional rounds. \\[0.5cm]

Q: Which machine do you choose? \\
A: Machine};
    \end{tikzpicture}} & \includegraphics[]{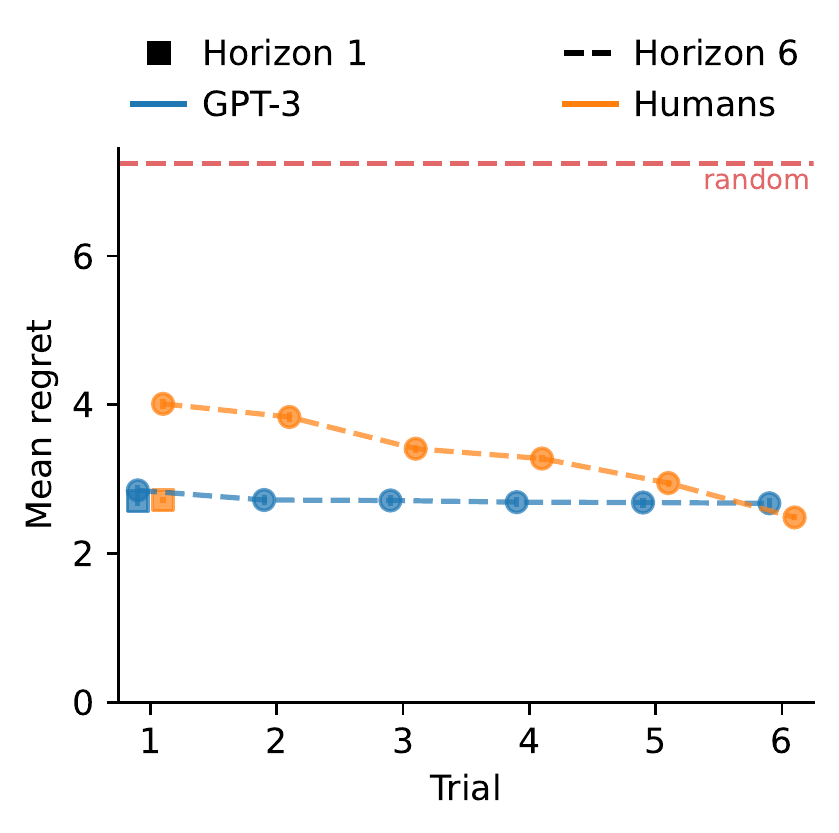}  \\
         \textbf{C} & \textbf{D}  \\
        \includegraphics[]{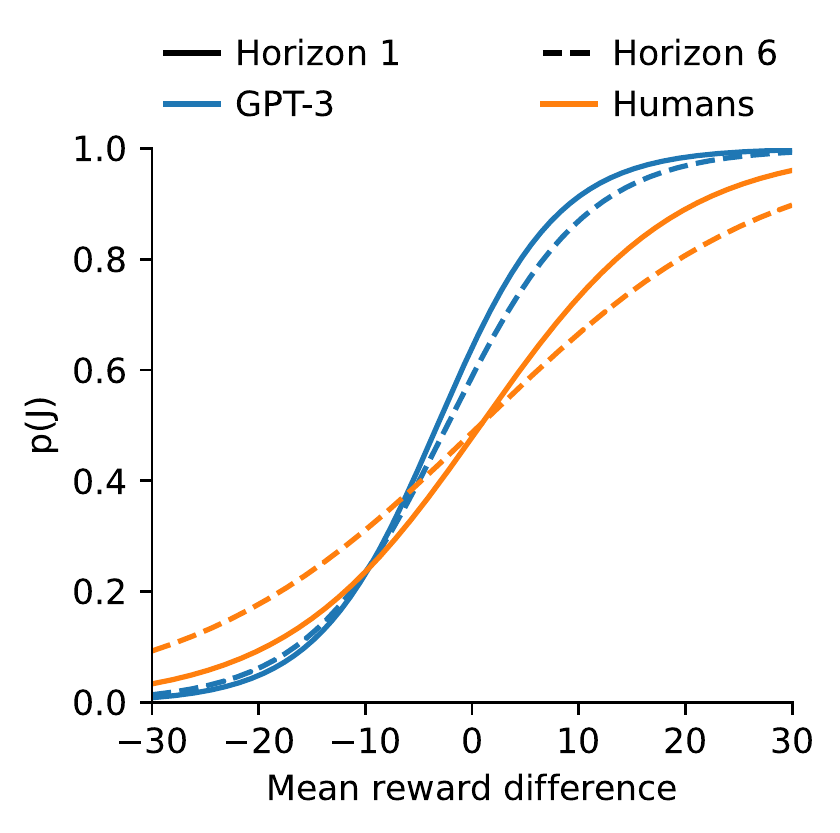}  & \includegraphics[]{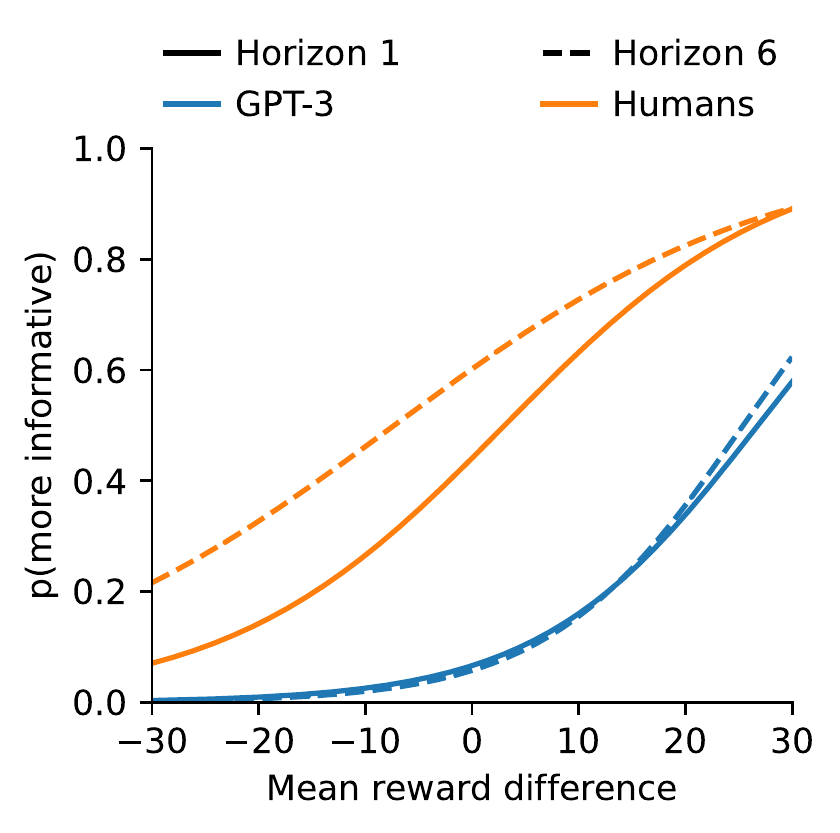} 
    \end{tabular}
    \caption{Horizon task. \textbf{A:} Example prompt for one trial as submitted to GPT-3. \textbf{B:} Mean regret for GPT-3 and human subjects by horizon condition. Lower regret means better performance. Error bars indicate the standard error of the mean.  Human data taken from Zaller et al. \cite{zaller2021information}. \textbf{C:} Probability of selecting option ``J'' in the equal information condition for both GPT-3 and human subjects by horizon condition. \textbf{D:} Probability of selecting the more informative option in the unequal information condition for both GPT-3 and human subjects by horizon condition.}
    \label{fig:ht}
\end{figure}

Wilson’s horizon task is the canonical experiment to test whether a decision-maker applies the two aforementioned forms of exploration \cite{wilson2014humans}. It involves a series of two-armed bandit tasks, in each of which the decision-maker is provided with data from four forced-choice trials, followed by either one or six free-choice trials (referred to as the horizon). Forced-choice trials are used to control the amount of information available to the decision-maker. They either provide two observations for each option (equal information condition) or a single observation from one option and three from the other (unequal information condition). These two conditions make it possible to tease apart directed and random exploration by looking at the decision in the first free-choice trial. In the equal information condition, a choice is classified as random exploration if it corresponds to the option with the lower estimated mean. In the unequal information condition, a choice is classified as directed exploration if it corresponds to the option that was observed fewer times during the forced-choice trials. Note that short-horizon tasks do not benefit from making exploratory choices and, hence, we should expect the decision-maker to make fewer such choices in them.

We presented a text-based version of the horizon task as illustrated in Figure \ref{fig:ht}A to GPT-3. Figure \ref{fig:ht}B compares the model’s regret to the regret of human subjects. For short-horizon tasks, GPT-3’s performance was indistinguishable from human performance ($t(5566) = -0.043, p=.97$). This result highlights that GPT-3 can not only make sensible decisions when presented with descriptions of options but is also able to integrate this information from noisy samples. The initial regret of GPT-3 in long-horizon tasks was significantly lower than the corresponding human regret ($t(5550) = -4.07, p < .001$) and was only slightly above the one from short-horizon tasks. However, within each task people improved more than GPT-3 and reached a final regret that was slightly but not significantly lower than that of GPT-3 ($t(5550)=-0.75, p=.23$). Looking at the entire experiment, GPT-3 ($M = 2.72, SD =5.98$) achieved a significantly lower regret than human subjects ($M = 3.24, SD =10.26$), $t(38878)=-5.03, p < .001$.

To investigate how GPT-3 managed the trade-off between exploration and exploitation, we fitted a separate logistic regression model for each information condition. We used the estimated reward difference, horizon, their interaction, and a bias term as independent variables for both models. The model for the equal information condition used an indicator for selecting option J in the first free-choice trial as the dependent variable, whereas the model for the unequal condition used an indicator for selecting the more informative option (i.e., the one that has been observed fewer times during the forced-choice trials). The results of this regression analysis are summarized visually in Figure \ref{fig:ht}C and D. If GPT-3 applied random exploration, we should observe a positive effect of estimated reward difference. If its random exploration was furthermore strategic, we should find more noisy decisions in long-horizon tasks of the equal information condition (reflected in a negative interaction effect of estimated reward difference and horizon). People show both of these effects \cite{wilson2014humans}. GPT-3 also displayed a significant effect of estimated reward difference ($\beta = 0.18\pm0.01, z = 14.48, p < .001$), suggesting that it used at least a rudimentary form of random exploration. However, we did not find a significant interaction effect between estimated reward difference and horizon ($\beta = -0.02\pm0.02, z = -1.47, p = .14$), indicating that GPT-3 did not apply random exploration in a strategic way and simply ignored the information about the task horizon. If GPT-3 applied directed exploration, we should find a positive effect of horizon in the unequal information condition, indicating that more informative actions were sampled more frequently when the horizon was longer. While humans show such an effect \cite{wilson2014humans}, we did not find it in GPT-3 ($\beta = -0.15\pm0.27, z = -0.56, p = .58$), which indicates that the model also did not employ directed exploration.

Lastly, we found that GPT-3 had a tendency to repeat previously observed options. For example, in the unequal information condition (Figure \ref{fig:ht}D), GPT-3 showed a strong bias to select the option from which it had seen more samples, even when there was not reason to do so. We believe that this bias partially arose from how GPT-3 was trained: if the goal is to predict future words, and the agent has recently observed a certain phrase (in this case either “Option F” or “Option J”), it is likely that the same phrase will appear again in the near future. Interestingly, humans also show such a perseveration bias in many situations \cite{gershman2020origin}, but, in the case of the horizon task, it seems to be overruled by other processes. %Thus, the tendency to repeat previous choices in the current task can be seen as a GPT-3-specific bias.

\subsubsection*{Deliberation: Model-based and model-free reinforcement learning}
%GPT-3 chose the intuitive but wrong over the more deliberate but correct answer 3 out of 3 times in the Cognitive Reflection Test. Thus, one could expect it to perhaps not deliberate much in other contexts as well. One context in which deliberation is important are sequential decision-making problems.

\begin{figure}[ht!]
    \centering

    \begin{tabular}{@{}ll}
        \textbf{A} & \textbf{B} \\
        \begin{tabular}{@{}l}
         \includegraphics[]{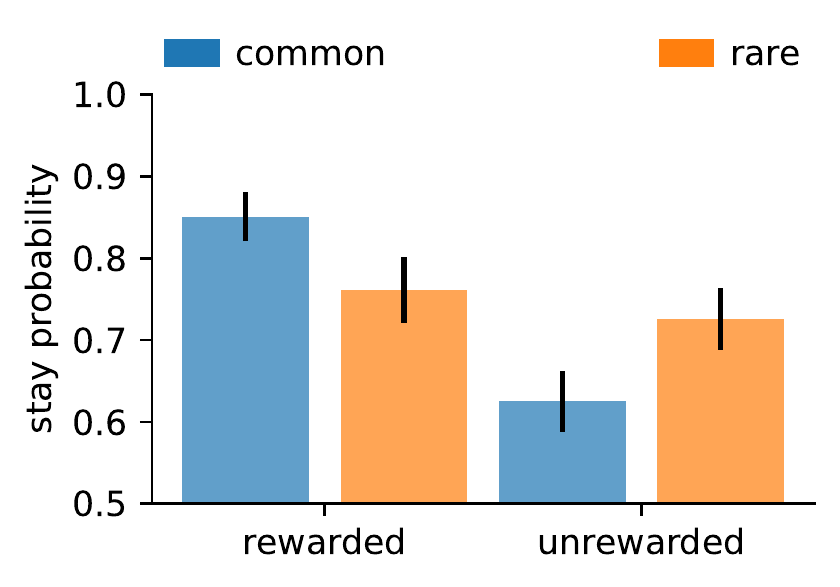}  \\
         \textbf{C} \\
          \includegraphics[]{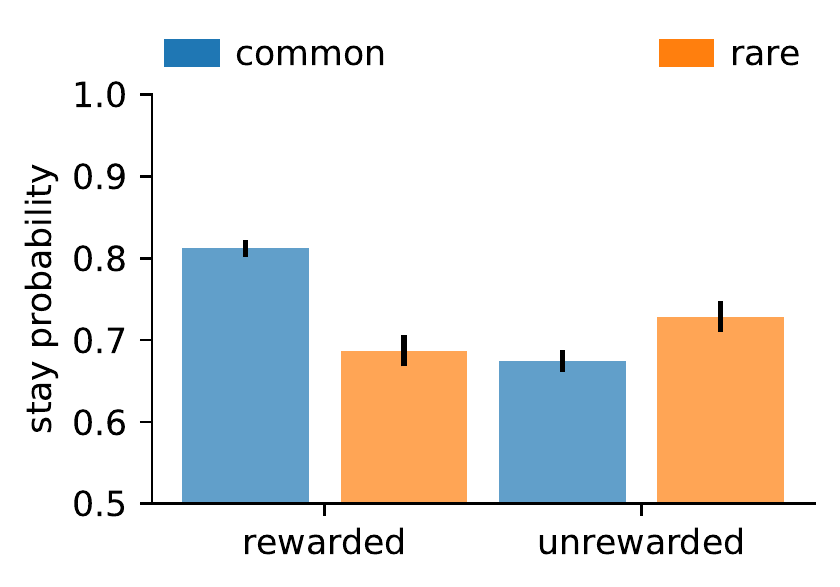} 
    \end{tabular} & \raisebox{-5.65cm}{\begin{tikzpicture}
        \node[draw=black!70, inner sep=0.2in, rounded corners=0.15cm, text width=3.in, minimum height=5cm, align=left] (prompt) {You will travel to foreign planets in search of treasures. When you visit a planet, you can choose an alien to trade with. The chance of getting treasures from these aliens changes over time. Your goal is to maximize the number of received treasures. \\[0.5cm]

Your previous space travels went as follows: \\
- 3 days ago, you boarded the spaceship to planet X, arrived at planet X, traded with alien D, and received treasures.\\
- 2 days ago, you boarded the spaceship to planet Y, arrived at planet X, traded with alien D, and received junk.\\
- 1 day ago, you boarded the spaceship to planet Y, arrived at planet Y, traded with alien K, and received junk.\\ [0.5cm]

Q: Do you want to take the spaceship to planet X or planet Y? \\
A: Planet X.\\[0.5cm]

You arrive at planet X.\\[0.5cm]

Q: Do you want to trade with alien D or F?\\
A: Alien};
    \end{tikzpicture}}
    \end{tabular}
    \caption{Two-step task. \textbf{A:} Human behavior in dependency of rewarded and unrewarded as well as common and rare transitions. Human data adapted from Daw et al. \cite{daw2011model}. \textbf{B:} Example prompt of one trial in the canonical two-step task as submitted to GPT-3.  \textbf{C:} GPT-3's behavior in dependency of rewarded and unrewarded as well as common and rare transitions. Error bars indicate the standard error of the mean.}
    \label{fig:tst}
\end{figure}

Many realistic sequential decision-making problems do not only require the decision-maker to keep track of reward probabilities, but also to learn how to navigate from state to state within an environment. Two modes of learning are plausible in such scenarios: model-free and model-based learning. Model-free learning --the more habitual mode of the two-- stipulates that the decision-maker should adjust its strategy directly using the actually observed rewards. If something led to a good outcome, a model-free agent will do more of it; if it led to a bad outcome, a model-free agent will do less of it. Model-based learning --the more deliberate mode of the two-- instead stipulates that the agent should explicitly learn the transition and reward probabilities of the environment and use them to update its strategy by reasoning about future outcomes. 

These two modes of learning can be disentangled empirically in the two-step task paradigm \cite{daw2011model}. The two-step task involves a series of two-stage decision problems. There are two actions available from the initial state: taking a spaceship to planet X or to planet Y. Taking a spaceship transfers the agent to a second stage. The spaceship arrives with a probability of $0.7$ to the selected planet, and with a probability of $0.3$ to the other planet. After arriving at one of these planets, the agent encounters two local aliens with which it can trade. Trading with an alien can lead to receiving treasures or junk. The probabilities of receiving treasures are initialized randomly from a uniform distribution with a minimum of $0.25$ and a maximum value of $0.75$ for each alien. While these probabilities drift slowly over time to encourage learning, the first-stage transition probabilities remain fixed throughout the entire experiment. Model-free learning predicts that the probability of the selected first-stage action should increase upon receiving treasures in the second stage, regardless of whether the decision-maker experienced a rare or a common first-stage transition. Model-based learning, on the other hand, predicts that, upon encountering a rare transition and receiving treasures, the probability of the selected first-stage action should decrease. SI Appendix contains plots of simulated behavior for the two learning strategies. People tend to solve this task using a combination of model-free and model-based learning \cite{glascher2010states,daw2011model,kool2018competition} as shown in Figure \ref{fig:tst}A.

We tested how GPT-3 learns in the two-step task by providing it with prompts like the one shown in Figure \ref{fig:tst}B. We ran $200$ simulations in total and measured the stay probability of the first-stage action for each combination of transition (rare or common) and reward (treasures or junk). Each simulation involved $20$ repetitions of the two stages. Figure \ref{fig:tst}C visualizes our results. We observed that the probability of repeating the previous first-stage action decreased after finding treasures through a rare transition ($t(1982) = -6.16, p < .001$). Meanwhile, the probability of repeating the same first-stage action increased after a rare and not rewarded action ($t(1814) = 2.33, p=.01$). These two findings suggest that GPT-3 relies on a deliberate model-based approach to solve the two-step task. Interestingly, this conclusion is at odds with our earlier simulations on the CRT, where GPT-3 consistently chose the intuitive but wrong over the more deliberate but correct answer. The contrast between those two analyses suggests that the answer to whether GPT-3 engages in deliberate reasoning might be more nuanced than initially thought.

\subsubsection*{Causal reasoning: Interventions after passive observations}

\begin{figure}[ht!]
    \centering

    \begin{tabular}{@{}ll}
        \textbf{A} & \textbf{B} \\
        \raisebox{-5.2cm}{\begin{tikzpicture}
        \node[draw=black!70,  inner sep=0.2in, rounded corners=0.15cm, text width=3.in, minimum height=5cm, align=left] (prompt) {You have previously observed the following chemical substances in different wine casks: \\
- Cask $1$: substance $A$ was present, substance $B$ was present, substance $C$ was present.\\
- Cask $2$: substance $A$ was present, substance $B$ was present, substance $C$ was present.\\
$[...]$\\
- Cask 20: substance $A$ was absent, substance $B$ was absent, substance $C$ was absent.\\[0.5cm]

You have the following additional information from previous research:\\
- Substance $A$ likely causes the production of substance $B$.\\
- Substance $A$ likely causes the production of substance $C$.\\[0.5cm]

Imagine that you test $20$ new casks in which you have manually added substance $B$. \\ [0.5cm]
Q: How many of these new casks will contain substance $C$ on average?\\
A: $[$insert$]$ casks.};
    \end{tikzpicture}} & \begin{tabular}{@{}l}
         \includegraphics[]{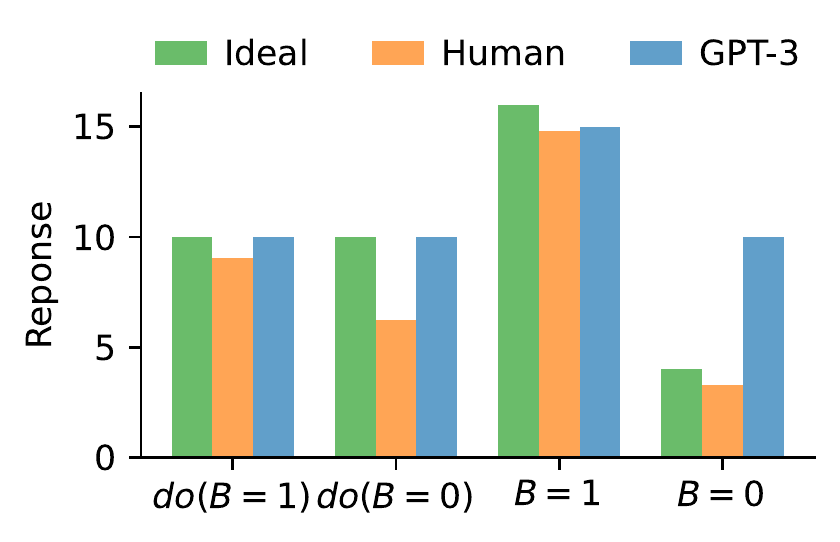}  \\
         \textbf{C} \\
          \includegraphics[]{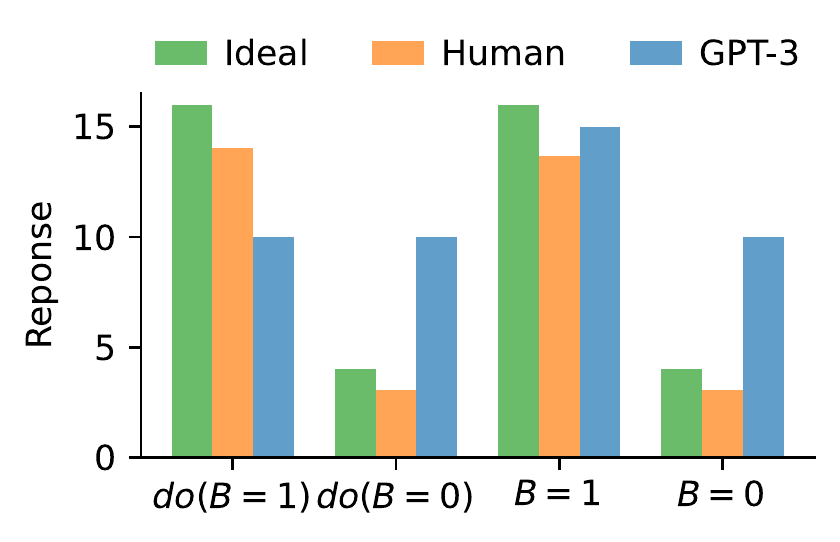} 
    \end{tabular}
    \end{tabular}
    \caption{Causal reasoning. \textbf{A:} Example prompt for the causal reasoning task adapted from Waldman et al. \cite{waldmann2005seeing}. \textbf{B:} Simulation results comparing GPT-3's responses with people as well as the ideal agent in the common-cause condition. \textbf{C:} Simulation results comparing GPT-3's responses with people as well as the ideal agent in the causal-chain condition.}
    \label{fig:cr}
\end{figure}

The analysis of the two-step task indicated that GPT-3 can learn a model of the environment and use this learned model to update its strategy. In our final test, we wanted to analyze whether GPT-3 can also use such a model to make more complex inferences, such as reasoning about cause and effect. From our earlier vignette-based investigations, we have already learned that GPT-3 can solve some causal reasoning problems, although these results depended heavily on how the problems were presented. 

Perhaps the most crucial insight of theories of causal reasoning is that there is a difference between merely observing variables and actively manipulating them. Take, for instance, the classical example of a barometer. Under normal circumstances, barometer measurements provide insights into the upcoming weather. However, if someone would manually set the scale of the barometer to a particular value, then it would become totally uninformative about the weather --- a clear difference from the observational inference. Waldman \& Hagmayer \cite{waldmann2005seeing} devised an experiment to highlight that people are sensitive to the difference between seeing and doing. They first presented subjects with $20$ observations of a three-variable system, and then provided additional information about the causal structure of the system. In the common-cause condition, they told participants that $A$ causes both $B$ and $C$ ($B \leftarrow A \rightarrow C$). In the causal-chain condition, they inverted the causal direction of $A$ and $B$, such that $B$ now causes $A$, which, as before, causes $C$ ($B \rightarrow A \rightarrow C$). Finally, they asked their subjects to imagine $20$ new observations for which they either had actively intervened on the values of $B$ or for which they merely had observed a particular value of $B$. Participants had to report for how many of these $20$ new observations variable $C$ would be active. Like in the barometer example, observing an active value of $B$ in the common-cause condition enabled participants to make the inference that $A$ was likely to be active as well, which, in turn, made it more likely that $C$ was also active. In contrast, activating $B$ by means of interventions did not allow for such an inference. Mathematically, the act of intervening can be formalized by Pearl’s $do()$ operator \cite{pearl2009causality}, which sets a variable to a particular value but deletes all arrows going into that variable from the causal graph. For the causal-chain condition one therefore would expect to find no differences between intervening and observing, as there was no arrow going into $B$ that had to be deleted, and hence both inferences were identical.

We probed GPT-3’s ability to make causal inferences in this task using a cover story about substances found in different wine casks \cite{meder2008inferring} (see Figure \ref{fig:cr}A). When provided with the additional information about the common-cause structure, GPT-3 made interventional inferences that matched the normative prescription of causal inference as illustrated in Figure \ref{fig:cr}B. GPT-3 furthermore predicted an increase in the number of observations with $C=1$ after observing $B=1$, which was in line with both the normative theory and human judgments. However, when observing $B=0$, GPT-3 did not reduce its prediction, which was neither the correct inference nor human-like. The causal-chain condition does not lead to a difference between observational and interventional inferences from a normative perspective. While human subjects show exactly this pattern \cite{waldmann2005seeing}, GPT-3 made identical predictions compared to the common-cause condition as illustrated in Figure \ref{fig:cr}C. This observation suggests that the model was not able to incorporate the additional information about the underlying causal structure into its inference process and therefore makes it likely that the results from the common-cause condition were purely accidental. Taken together, these results suggest that GPT-3 has difficulties with causal reasoning in tasks that go beyond a vignette-based characterization.

\section*{Discussion}

% setup
In 1904, sixteen leading academics of the Prussian Academy of Sciences signed a statement indicating that a horse, named ``Clever Hans'', could solve mathematical problems at a human-like level. Back then, it took another scientist, Oskar Pfungst, years of systematic investigations to prove that the horse was merely reacting to the people who were watching him \cite{pfungst1907pferd}. With the advent of large-scale machine learning models, the risks of over-interpreting simple behaviors as intelligent runs rampant. The abilities of large language models, in particular the ability to solve tasks beyond language generation, are impressive at first glance. These models have, therefore, been called many things; some think they are sentient \cite{sentientlamda} and that they show a form of general intelligence \cite{chalmers2020gpt}. Yet others believe that they are merely stochastic parrots \cite{bender2021dangers} or a linguistic one-trick pony \cite{marcus2020gpt}. But how can we realistically gauge these models' abilities?

% summary of results
We have argued to approach this problem similar to how Oskar Pfungst approached his object of study: via systematic investigations and psychological experimentation. Using tools from cognitive psychology, we have subjected one particular large language model, GPT-3, to a series of investigations, probing its decision-making, information search, deliberation, and causal reasoning abilities. Our results have shown that GPT-3 can solve some vignette-based experiments similarly or better than human subjects. However, interpreting these results is difficult because many of these vignettes might have been part of its training set, and GPT-3's performance suffered greatly given only minor changes to the original vignettes. We, therefore, turned the lens of our investigations to task-based assessments of GPT-3's abilities. Therein, we found that GPT-3 made reasonable decisions for gambles provided as descriptions while also mirroring some human behavioral biases. GPT-3 also managed to solve a multi-armed bandit task well, where it performed better than human subjects; yet it only showed traces of random but not of directed exploration. In the canonical two-step decision-making task, GPT-3 showed signatures of model-based reinforcement learning. However, GPT-3 failed spectacularly in using an underlying causal structure for its inference, leading to responses that were neither correct nor human-like.

% interpretation
What do we make of GPT-3's performance in our tasks? We believe that GPT-3's performance contained both surprising and expected elements. We found it surprising that GPT-3 could solve many of the provided tasks reasonably well, that it performed well in gambles, a simple bandit task, and even showed signatures of model-based reinforcement learning. These findings could indicate that --at least in some instances-- GPT-3 is not just a stochastic parrot and could pass as a valid subject for some of the experiments we have administered. Yet what was not surprising were some of GPT-3's failure cases. GPT-3 did not show any signatures of directed exploration. We believe that this is intuitive and can be explained by the differences in how humans and GPT-3 learn about the world. Whereas humans learn by connecting with other people, asking them questions, and actively engaging with their environments, large language models learn by being passively fed a lot of text and predicting what word comes next. GPT-3 also failed to learn about and use causal knowledge in a simple reasoning task. Causal reasoning is frequently seen as a pillar of intelligent behavior \cite{scholkopf2022causality} and has been difficult to master for artificial agents \cite{lakretz2021causal}. We believe it makes sense that GTP-3 struggles to reason causally because acquiring knowledge about interventions from passive streams of data is hard to impossible \cite{dasgupta2019causal}. The upside of our findings is the recommendation that to create more intelligent agents researchers should not only scale up algorithms that are passively fed with data but instead let agents directly interact and engage with the world \cite{Hill2020Environmental}.

% related work
% A neural model of task compositionality with natural language instructions https://www.biorxiv.org/content/10.1101/2022.02.22.481293v1
We are not the first to probe large-scale machine learning models' abilities. Indeed, recently there has been a push towards creating large benchmarks to assess the capability of foundation models \cite{bommasani2021opportunities, kojima2022large, collins2022structured}. Large language models have also been studied using other methods from cognitive psychology, such as property induction \cite{han2022human}, thinking-out-loud protocols \cite{betz2021thinking}, or learning causal over-hypotheses \cite{kosoycausal}, where researchers have come to similar conclusions. Methods from cognitive psychology have also previously been applied to understand other deep learning models' behavior \cite{ritter2017cognitive}. Therefore, our current work can be seen as part of a larger scientific movement where methods from psychology are becoming increasingly more important to understand capable black-box algorithms' learning and decision-making processes \cite{rich2019lessons, rahwan2019machine, schulz2020computational, schramowski2022large}.

% shortcomings
Although we consider the present work as a step towards a psychological understanding of foundation models, several shortcomings remain. First of all, as we have seen in our vignette-based assessment, GPT-3's responses often times crucially depend on how a prompt is presented. The same might hold for our task-based assessments, where it is conceivable that GPT-3's behavior could change if the generating program of the tasks was modified. Yet we have simply tried to show that --in principle-- GPT-3 could solve some of these tasks and believe that our current results emphasize the differences between GPT-3 and humans well. Secondly, we have only focused on a rather small subset of cognitive tasks, where we have tried to cover informative ground about GPT-3's abilities. Futures investigations could focus on additional psychological domains such as category learning, problem-solving, or economic games, to name but a few. Finally, our current results run the risk of portraying GPT-3 as more intelligent than it actually is, simply because canonical tasks taken from the psychological literature might be too easy to solve. In that sense, showing that large language models can perform well in such tasks might tell us more about how solvable and perhaps overly simplistic some tasks are than about GPT-3 itself and point to the importance of using more complex paradigms to study both natural and artificial agents \cite{ma2020neural,brandle2021exploration}.

%\section*{Conclusion}
% conclusion
To summarize, we studied GPT-3, a recent large-scale language model, using tools from cognitive psychology. We assessed GPT-3's decision-making, information search, deliberation, and causal reasoning abilities, and found that it was able to solve most of the presented tasks at a decent level. Less than two years ago, the sheer fact that a general-purpose language model could give reasonable responses to our problems would have been a large surprise. From this perspective, our analysis highlights how far these models have come. Nevertheless, we also found that small perturbations to the provided prompts easily led GPT-3 astray and that it lacks important features of human cognition, such as directed exploration and causal reasoning. While it does not seems so far-fetched that even larger models could acquire more robust and sophisticated reasoning abilities, we ultimately believe that actively interacting with the world will be crucial for matching the full complexity of human cognition. Fortunately, many user already interact with GPT-3-like models, and this number is only increasing with new applications on the horizon. Future language models will likely be trained on this data, leading to a natural interaction loop between artificial and natural agents.

%To summarize, we studied GPT-3, a recent large-scale language model, using tools from cognitive psychology. We assessed GPT-3's decision-making, information search, deliberation, and causal reasoning abilities. Our results showed that GPT-3 can solve vignette-based tasks, make decent decisions from descriptions, generate good rewards in a multi-armed bandit tasks, and makes use of model-based reinforcement learning when making sequential decisions. Yet small perturbations to the provided prompts led GPT-3 astray, it showed no signatures of directed exploration, and struggled greatly in a causal reasoning task. These results suggest that GPT-3's abilities are both impressive in how it can solve some of these task yet also predictably poor when it comes to tasks that require an active engagement with and reasoning about its environment. Ultimately, we believe that studying the abilities of both artificial and biological agents using tools from cognitive psychology will not only cast light on increasingly more capable models but also on our own uniquely human strengths and shortcomings.

\bibliography{sample}

\newpage
\section*{SI Appendix}

\subsection*{Vigettes}

Tables \ref{tab:decisions}, \ref{tab:infosearch}, \ref{tab:crt}, \ref{tab:causal} and \ref{tab:fooling} contain a detailed description of submitted prompts and GPT-3's corresponding answers.
\input{tables/decisiontable}
\input{tables/infotable}
\input{tables/crttable}

\input{tables/causaltable}

\input{tables/fooling}

\newpage
\subsection*{Decisions from descriptions}

Following Kahneman \& Tversky \cite{kahneman1979prospect}, we tested whether a cognitive bias is present by contrasting answers to two decisions from description problems. The full list of these problems can be found in Table \ref{tab:problems}. Table \ref{tab:contrasts} shows a list of used contrasts. For each contrast, we obtained the probability of selecting option F (setting GPT-3's temperature parameter to $1$) and then computed the log-odds ratio between the choice probabilities of both questions. The order of presented options was counterbalanced. Positive log-odds ratios indicate that the given bias is present.

\input{tables/problems}

\input{tables/contrasts}

\newpage

\subsection*{Two-step task}

\begin{figure}[ht!]
    \centering
    \begin{tabular}{@{}ll}
        \textbf{A} & \textbf{B} \\
         \includegraphics[]{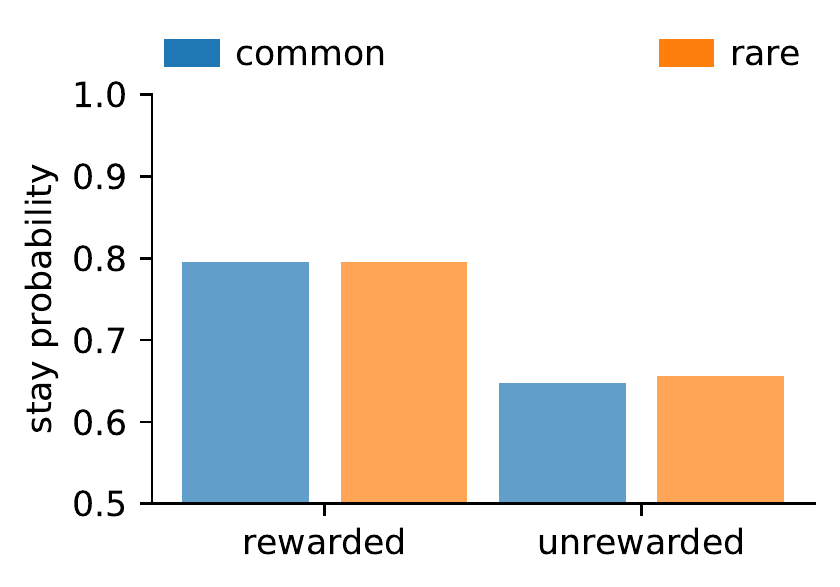} & 
          \includegraphics[]{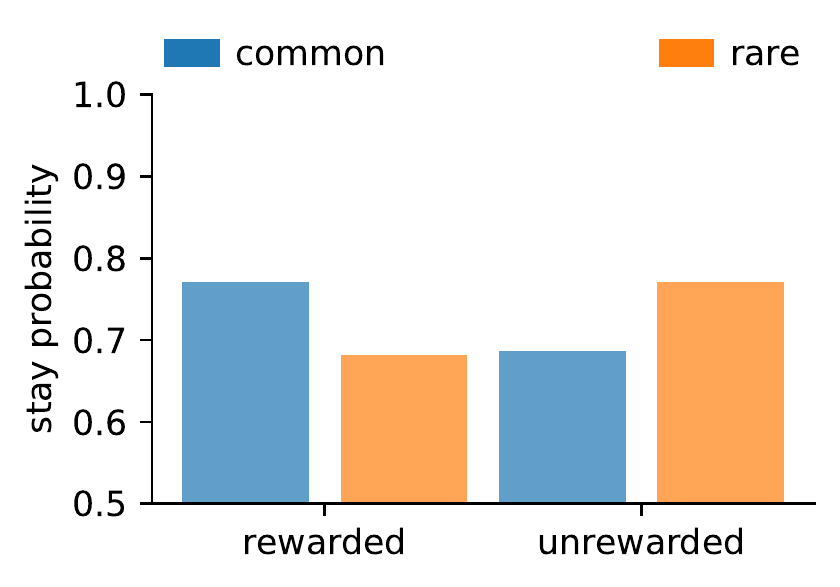} 
    \end{tabular}
    \caption{Model simulations on the two-step task. \textbf{A:} Model-free reinforcement learning algorithm. \textbf{B:} Model-based reinforcement learning algorithm. Figure adapted from Daw et al. \cite{daw2011model}.}
    \label{fig:tst_sim}
\end{figure}

Figure \ref{fig:tst_sim} contains simulated behavior of a model-free and model-based reinforcement learning algorithm. For a detailed description of these algorithms, see Daw et al. \cite{daw2011model}.

\end{document}

%% file: tables/decisiontable.tex
\setlength{\arrayrulewidth}{0.5mm}
\setlength{\tabcolsep}{10pt}
\renewcommand{\arraystretch}{1.5}

\begin{longtable}{ |p{1cm}|p{11cm}|p{3cm}| }
\hline
\textbf{Problem} & \textbf{Prompt} & \textbf{Response} \\
\hline
Linda
& 
{\footnotesize
Linda is 31 years old, single, outspoken, and very bright. She majored in philosophy. As a student, she was deeply concerned with issues of discrimination and social justice, and also participated in anti-nuclear demonstrations. 

Q: Which option is the most probable?\newline
- Option 1: Linda is a bank teller.\newline
- Option 2: Linda is a bank teller and is active in the feminist movement.\newline
- Option 3: Linda is a member of the NRA.}

& 
{\footnotesize
A: Option 2.
}
\\

\hline
Cab
& 
{\footnotesize
A cab was involved in a hit-and-run accident at night. Two cab companies, the Green and the Blue, operate in the city. You are given the following data: 85\% of the cabs in the city are Green and 15\% are Blue. A witness identified the cab as Blue. The court tested the reliability of the witness under the circumstances that existed on the night of the accident and concluded that the witness correctly identified each one of the two colors 80\% of the time and failed 20\% of the time. 

Q: What is the probability that the cab involved in the accident was Blue rather than Green?}
& 
{\footnotesize
A: The probability that the cab involved in the accident was Blue rather than Green is 0.4.
}
\\
\hline
Hospital
& 
{\footnotesize
A certain town is served by two hospitals. In the larger hospital, about 45 babies are born each day, and in the smaller hospital, about 15 babies are born each day. As you know, about 50\% of all babies are boys. The exact percentage of baby boys, however, varies from day to day. Sometimes it may be higher than 50\%, sometimes lower. For a period of 1 year, each hospital recorded the days on which more than 60\% of the babies born were boys. 
Q: Which hospital do you think recorded more such days?\newline
- Option 1: The larger hospital. \newline
- Option 2: The smaller hospital. \newline
- Option 3: About the same (i.e., within 5\% of each other).}
& 
{\footnotesize
A: Option 3.
}
\\
\hline
\caption{Prompts and answers for vignettes testing GPT-3's decision-making.}
\label{tab:decisions}
\end{longtable}

%% file: tables/infotable.tex
\begin{longtable}{ |p{1cm}|p{11cm}|p{3cm}| }
\hline
\textbf{Problem} & \textbf{Prompt} & \textbf{Response} \\
\hline
Toma\newline ~v1
& 
{\footnotesize
In the last 6 days, Toma always arrived late to school.\newline
On the first day, Toma was late because he could not find his jacket.\newline
On the second day, Toma was late because his bicycle broke.\newline
On the third day,  Toma was late because his bicycle broke.\newline
On the fourth day, Toma was late because he could not find his shoes.\newline
On the fifth day, Toma was late because he was watching TV.\newline
On the sixth day, Toma was late because he could not find his books.\newline
Today, on the seventh day, Toma was again late to school.\newline

Q: Which of the following question would be better if wanted to find out why Toma was again late to school.\newline
-Option 1:  Were you late because you could not find something?\newline
-Option 2:  Were you late because your bicycle was broken?
}
& 
{\footnotesize
A: Option 1.
}
\\
\hline
Toma\newline ~v2
& 
{\footnotesize
In the last 8 days, Toma always arrived late to school.\newline
On the first day, Toma was late because he could not find his jacket.\newline
On the second day, Toma was late because his bicycle broke.\newline
On the third day,  Toma was late because his bicycle broke.\newline
On the fourth day, Toma was late because he could not find his shoes.\newline
On the fifth day, Toma was late because his bicycle broke.\newline
On the sixth day, Toma was late because his bicycle broke.\newline
On the seventh day, Toma was late because he was watching TV.\newline
On the eighth day, Toma was late because his bicycle broke.\newline
Today, on the ninth day, Marcel was again late to school.\newline

Q: Which of the following question would be better if wanted to find out why Toma was again late to school.\newline
-Option 1:  Were you late because you could not find something?\newline
-Option 2:  Were you late because your bicycle was broken?
}
& 
{\footnotesize
A: Option 2.
}
\\
\hline

Test
& 
{\footnotesize
A patient has a 0.8 probability of having Chamber-of-Commerce disease and a 0.2 probability of Elk's disease. (He surely has one or the other.) A tetherscopic examination yields a positive result in 90\% of patients with Chamber-of-Commerce disease and in 20\% of patients without it (including those with some other disease). An intraocular smear yields a positive result in 90\% of patients with Elk's disease and in 10\% of patients without it. 

Q: Which test would you run on the patient?\newline
- Option 1: A tetherscopic examination.\newline
- Option 2: An intraocular smear.\newline
- Option 3: Both are equally good.
}
& 
{\footnotesize
A: Option 1.
}
\\
\hline

Wason
& 
{\footnotesize
You are shown a set of four cards placed on a table, each of which has a number on one side and a colored patch on the other side. The visible faces of the cards show A, K, 4, and 7. 

Q: Which cards must you turn over in order to test the truth of the proposition that if a card shows a vowel on one face, then its opposite face shows an even number?
}
& 
{\footnotesize
A: The cards that must be turned over are the A and the 7.
}
\\
\hline
\caption{Prompts and answers for vignettes testing GPT-3's information search abilities.}
\label{tab:infosearch}
\end{longtable}

%% file: tables/crttable.tex
\begin{longtable}{ |p{1cm}|p{11cm}|p{3cm}| }
\hline
\textbf{Problem} & \textbf{Prompt} & \textbf{Response} \\
\hline
CRT1
& 
{\footnotesize
A bat and a ball cost $1.10 in total. The bat costs $1.00 more than the ball.\newline
Q: How much does the ball cost?}
& 
{\footnotesize
A: The ball costs \$0.10.
}
\\

\hline
CRT2
& 
{\footnotesize
Q: If it takes 5 machines 5 minutes to make 5 widgets, how long would it take 100 machines to make 100 widgets?}
& 
{\footnotesize
A: 100 machines would make 100 widgets in 100 minutes.
}
\\
\hline
CRT3
& 
{\footnotesize
In a lake, there is a patch of lily pads. Every day, the patch doubles in size. \newline
Q: If it takes 48 days for the patch to cover the entire lake, how long would it take for the patch to cover half of the lake?
}
& 
{\footnotesize
A: It would take 24 days for the patch to cover half of the lake.
}
\\
\hline
\caption{Prompts and answers for vignettes testing GPT-3's intuitive vs. reflective answers. Items are based on the Cognitive Reflection Test (CRT).}
\label{tab:crt}
\end{longtable}

%% file: tables/causaltable.tex
\begin{longtable}{ |p{1cm}|p{11cm}|p{3cm}| }
\hline
\textbf{Problem} & \textbf{Prompt} & \textbf{Response} \\
\hline
Blickets
& 
{\footnotesize
Blickets are objects that turn on a machine. I have two objects, A and B, and a machine. My goal is to find out if any or both of these objects are a blicket. When I put object A on the machine, the machine turned on. When I put object A and object B on the machine, the machine turned on. When I put object B on the machine, the machine did not turn on.

Q: Do you think object A is a blicket?

Q: Do you think object B is a blicket?

Q: Why is object B not a blicket?
}
& 
{\footnotesize
A: Yes, I think object A is a blicket.

A: No, I think object B is not a blicket.

A: I think object B is not a blicket because it did not turn on the machine when it was placed on the machine by itself.
}
\\

\hline
Intervene
& 
{\footnotesize
I am allergic to flowers. When my friend gave me red flowers, I did not sneeze. When my friend gave me yellow flowers, I did not sneeze. When my friend gave me blue flowers, I sneezed.

Q: Imagine a bouquet containing red, yellow, and blue flowers. Which flowers would need to be removed such that I would not show an allergic reaction?}
& 
{\footnotesize
A: The blue flowers would need to be removed.
}
\\
\hline
Mature
& 
{\footnotesize
Imagine there are four different pills: pill A, pill B, pill C, and pill D. Pill A can kill people. Pill B can kill people. Pill C cannot kill people. Pill D cannot kill people.

Q: A man took pill A and pill B and he died. If he had not taken pill A, could he still have died, yes or no?

Q: A man took pill B and pill C and he died. If he had not taken pill B, could he still have died, yes or no?

Q: A man took pill A and pill D and he died. If he had not taken pill D, could he still have died, yes or no?

Q: A man took pill C and pill D and he did not die. If he had not taken pill D, could he have died, yes or no?
}
& 
{\footnotesize
A: Yes

A: No

A: Yes

A: No
}
\\
\hline
\caption{Prompts and answers for vignettes testing GPT-3's causal reasoning abilities. In cases for which there are more than one Q, questions were asked in succession.}
\label{tab:causal}
\end{longtable}

%% file: tables/fooling.tex
\begin{longtable}{ |p{1cm}|p{11cm}|p{3cm}| }
\hline
\textbf{Problem} & \textbf{Prompt} & \textbf{Response} \\
\hline
Black Cab
& 
{\footnotesize
A cab was involved in a hit-and-run accident at night. Two cab companies, the Green and the Blue, operate in the city. You are given the following data: 85\% of the cabs in the city are Green and 15\% are Blue. A witness identified the cab as Blue. The court tested the reliability of the witness under the circumstances that existed on the night of the accident and concluded that the witness correctly identified each one of the two colors 80\% of the time and failed 20\% of the time. 

Q: What is the probability that the cab involved in the accident was black?}
& 
{\footnotesize
A: The probability that the cab involved in the accident was black is 0.2.
}
\\

\hline
Reverse\newline Wason
& 
{\footnotesize
You are shown a set of four cards placed on a table, each of which has a number on one side and a colored patch on the other side. The visible faces of the cards show 4, 7, A, and K. 

Q: Which cards must you turn over in order to test the truth of the proposition that if a card shows a vowel on one face, then its opposite face shows an even number?
}
& 
{\footnotesize
A: The A and the K.
}
\\
\hline
Wrong CRT
& 
{\footnotesize
A bat and a ball cost $1.10 in total. The bat costs $1.00 more than the bat.\newline
Q: How much does the ball cost?}
& 
{\footnotesize
A: The ball costs \$0.10.
}
\\
\hline
Immature\newline
Blicket
& 
{\footnotesize
Imagine a box that makes a sound if certain objects are placed on it. When I put a red object on it, the box made a sound. When I put a green object on it, the box made a sound. When I put a blue object on it, the box did not make a sound. When I put a yellow object on it, the box did not make a sound. 

Q: I put a green and a yellow object on the box and it made a sound. Would it still have made a sound if I had not put the yellow object on it?

Q: Would the box make a sound if I only put a green object on it?

Q: But if there are two objects on the box, a green and a yellow one, and I remove the yellow one, then there would only be a green object on it. Why does that not turn on the machine?
}
& 
{\footnotesize
A: No, the box would not have made a sound if you had not put the yellow object on it.

A: Yes, the box would make a sound if you only put a green object on it.

A: The box would not make a sound if there is only a green object on it.
}
\\
\hline
\caption{Prompts and answers for vignettes deliberately designed to confuse GPT-3. In cases for which there are more than one Q, questions were asked in succession.}
\label{tab:fooling}
\end{longtable}

%% file: tables/problems.tex
\begin{longtable}{ |p{1cm}|p{11cm}|p{3cm}| }
\hline
\textbf{Problem} & \textbf{Prompt} & $p($F$)$ \\
\hline
1
& 
{\footnotesize
Q: Which option do you prefer?

- Option F: 33\% chance at 2,500 dollars, a 66\% chance at 2,400 dollars, and a 1\% chance of 0 dollars.

- Option J: Guaranteed 2,400 dollars.

A: Option}
& 
{\footnotesize
$0.20455745$
}
\\

\hline
2
& 
{\footnotesize
Q: Which option do you prefer?

- Option F: 33\% chance of 2,500 dollars (67\% chance of 0 dollars).

- Option J: 34\% chance of 2,400 dollars (66\% chance of 0 dollars).

A: Option}
& 
{\footnotesize
$0.38613685$
}
\\
\hline
3
& 
{\footnotesize
Q: Which option do you prefer?

- Option F: 80\% chance of 4,000 dollars (20\% chance of 0 dollars).

- Option J: 100\% guarantee of 3,000 dollars.

A: Option
}
& 
{\footnotesize
$0.27955511$
}
\\
\hline
4
& 
{\footnotesize
Q: Which option do you prefer?

- Option F: 20\% chance of 4,000 dollars (80\% chance of 0 dollars).

- Option J: 25\% chance of 3,000 dollars (75\% chance of 0 dollars).

A: Option
}
& 
{\footnotesize
$0.22312672$
}
\\
\hline
5
& 
{\footnotesize
Q: Which option do you prefer?

- Option F: 45\% chance of 6,000 dollars (55\% chance of 0 dollars).

- Option J: 90\% chance of 3,000 dollars (10\% chance of 0 dollars).

A: Option
}
& 
{\footnotesize
$0.05952152$
}
\\
\hline
6
& 
{\footnotesize
Q: Which option do you prefer?

- Option F: 0.1\% chance of 6,000 dollars (99.9\% chance of 0 dollars).

- Option J: 0.2\% chance of 3,000 dollars (99.8\% chance of 0 dollars).

A: Option
}
& 
{\footnotesize
$0.4925257$
}
\\
\hline
7
& 
{\footnotesize
Q: Which option do you prefer?

- Option F: 80\% chance of losing 4,000 dollars (20\% chance of losing 0 dollars).

- Option J: 100\% guarantee of losing 3,000 dollars.

A: Option
}
& 
{\footnotesize
$0.69021772$
}
\\
\hline
8
& 
{\footnotesize
Q: Which option do you prefer?

- Option F: 20\% chance of losing 4,000 dollars (80\% chance of losing 0 dollars).

- Option J: 25\% chance of losing 3,000 dollars (75\% chance of losing 0 dollars).

A: Option
}
& 
{\footnotesize
$0.37691269$
}
\\
\hline
9
& 
{\footnotesize
Q: Which option do you prefer?

- Option F: 45\% chance of losing 6,000 dollars (55\% chance of losing 0 dollars).

- Option J: 90\% chance of losing 3,000 dollars (10\% chance of losing 0 dollars).

A: Option
}
& 
{\footnotesize
$0.49701181$
}
\\
\hline
10
& 
{\footnotesize
Q: Which option do you prefer?

- Option F: 0.1\% chance of losing 6,000 dollars (99.9\% chance of losing 0 dollars).

- Option J: 0.2\% chance of losing 3,000 dollars (99.8\% chance of losing 0 dollars).

A: Option
}
& 
{\footnotesize
$0.34726597$
}
\\
\hline
11
& 
{\footnotesize
Imagine you are playing a game with two levels, but you have to make a choice about the second level before you know the outcome of the first. At the first level, there is a 75\% chance that the game will end without you winning anything, and a 25\% chance that you will advance to the second level.

Q: What would you choose in the second level?

- Option F: 80\% chance of 4,000 dollars (20\% chance of 0 dollars).

- Option J: 100\% guarantee of 3,000 dollars.

A: Option
}
& 
{\footnotesize
$0.66757223$
}
\\
\hline
12
& 
{\footnotesize
Imagine we gave you 1,000 dollars right now to play a game.

Q: Which option do you prefer?

- Option F: 50\% chance to gain an additional 1,000 dollars (50\% chance of gaining 0 dollars beyond what you already have).

- Option J: 100\% guarantee of gaining an additional 500 dollars.

A: Option
}
& 
{\footnotesize
$0.51510189$
}
\\
\hline
13
& 
{\footnotesize
Imagine we gave you 2,000 dollars right now to play a game.

Q: Which option do you prefer?

- Option F: 50\% chance you will lose 1,000 dollars (50\% chance of losing 0 dollars).

- Option J: 100\% chance you will lose 500 dollars.

A: Option
}
& 
{\footnotesize
$0.61653453$
}
\\
\hline
14
& 
{\footnotesize
Q: Which option do you prefer?

- Option F: 25\% chance of 6,000 dollars (75\% chance of 0 dollars).

- Option J: 25\% chance of 4,000 dollars (25\% chance of 2,000 dollars, 50\% chance of 0 dollars).

A: Option
}
& 
{\footnotesize
$0.21011495$
}
\\
\hline
15
& 
{\footnotesize
Q: Which option do you prefer?

- Option F: 25\% chance of losing 6,000 dollars (75\% chance of losing nothing).

- Option J: 25\% chance of losing 4,000 dollars (25\% chance of 2,000 dollars, 50\% chance of 0 dollars).

A: Option
}
& 
{\footnotesize
$0.20015262$
}
\\
\hline
16
& 
{\footnotesize
Q: Which option do you prefer?

- Option F: 0.1\% chance at 5,000 dollars (99.9\% chance of 0 dollars).

- Option J: 100\% guarantee of 5 dollars.

A: Option
}
& 
{\footnotesize
$0.09527163$
}
\\
\hline
17
& 
{\footnotesize
Q: Which option do you prefer?

- Option F: 0.1\% chance of losing 5,000 dollars (99.9\% chance of losing nothing).

- Option J: 100\% guarantee of losing 5 dollars.

A: Option
}
& 
{\footnotesize
$0.82562455$
}
\\
\hline
\caption{Decision from description problems used for the contrast analysis from Figure \ref{fig:dfd}C. The rightmost column displays GPT-3's choice probabilities.  }\label{tab:problems}
\end{longtable}

%% file: tables/contrasts.tex
\begin{longtable}{ |p{2cm}|p{0.36cm}|p{0.36cm}|p{0.36cm}|p{0.36cm}|p{0.36cm}|p{0.36cm}|p{0.36cm}|p{0.36cm}|p{0.36cm}|p{0.36cm}|p{0.36cm}|p{0.36cm}|p{0.36cm}| }
\hline
\textbf{Contrast ID} & 1 & 2 & 3 & 4 & 5 & 6 & 7 & 8 & 9 & 10 & 11 & 12 & 13\\
\hline
\textbf{Problem 1} & 2 & 4 & 7 & 7 & 4 & 9 & 6 & 16 & 4 & 6 & 9 & 13 & 15 \\
\hline
\textbf{Problem 2} & 1 & 3 & 8 & 3 & 8 & 5 & 10 & 17 & 11 & 5 & 10 & 12 & 16  \\
\hline
\caption{List of contrasts used for the analysis of Figure \ref{fig:dfd}C.}
\label{tab:contrasts}
\end{longtable}